\documentclass[sigconf,nonacm]{acmart}

\AtBeginDocument{}

\usepackage{amsmath}
\usepackage{algorithm}
\usepackage{algpseudocode}
\usepackage{booktabs}
\usepackage{array}

\microtypesetup{protrusion=false}
\setcopyright{none}
\acmDOI{}
\acmISBN{}
\renewcommand\footnotetextcopyrightpermission[1]{}

\begin{document}

\title[PersonaAgent: Retrieval-Augmented Transaction Anomaly Scoring]{PersonaAgent:
Retrieval-Augmented Transaction Anomaly Scoring with Refreshable Multi-Entity
Persona Memory}

\author{Zhengyu Fang}
\correspondingauthor
\affiliation{%
  \institution{Case Western Reserve University}
  \city{Cleveland}
  \state{OH}
  \country{USA}}
\email{zxf177@case.edu}

\author{Liang Wang}
\affiliation{%
  \institution{Visa Research}
  \city{Foster City}
  \state{CA}
  \country{USA}}
\email{liawang@visa.com}

\author{Yingtong Dou}
\affiliation{%
  \institution{Visa Research}
  \city{Foster City}
  \state{CA}
  \country{USA}}
\email{yidou@visa.com}

\author{Yuzhong Chen}
\affiliation{%
  \institution{Visa Research}
  \city{Foster City}
  \state{CA}
  \country{USA}}
\email{yuzchen@visa.com}

\author{Shubham Jain}
\affiliation{%
  \institution{Visa Research}
  \city{Foster City}
  \state{CA}
  \country{USA}}
\email{shubhjai@visa.com}

\author{Junpeng Wang}
\affiliation{%
  \institution{Visa Research}
  \city{Foster City}
  \state{CA}
  \country{USA}}
\email{junpenwa@visa.com}

\author{Xiran Fan}
\affiliation{%
  \institution{Visa Research}
  \city{Foster City}
  \state{CA}
  \country{USA}}
\email{xirafan@visa.com}

\author{Menghai Pan}
\affiliation{%
  \institution{Visa Research}
  \city{Foster City}
  \state{CA}
  \country{USA}}
\email{menpan@visa.com}

\author{Jiarui Sun}
\affiliation{%
  \institution{Visa Research}
  \city{Foster City}
  \state{CA}
  \country{USA}}
\email{jiaruis2@visa.com}

\author{Yan Zheng}
\affiliation{%
  \institution{Visa Research}
  \city{Foster City}
  \state{CA}
  \country{USA}}
\email{yazheng@visa.com}

\author{Yiwei Cai}
\affiliation{%
  \institution{Visa Research}
  \city{Austin}
  \state{TX}
  \country{USA}}
\email{yicai@visa.com}

\renewcommand{\shortauthors}{Fang et al.}

\begin{abstract}
In large-scale payment networks, entities (card, merchant, etc.) and their
interactions play a key role in transaction anomaly detection. However,
transaction patterns between different entities usually change more frequently
than the production model refresh schedule. Meanwhile, entity embedding tables
suffer from high-cardinality and sparse interaction issues. To
resolve the above challenges, we present \textbf{\mbox{PersonaAgent}}, an agentic
transaction anomaly detection framework which represents the latest entity
behavior signals as personas and saves them in a refreshable memory. Offline, a
weekly scheduled batch job is used to extract the latest behavior signals and
update them in the memory. At inference time, a lightweight adapter retrieves the
corresponding entity behavior signals of each transaction from memory and uses
them to augment the base anomaly classifier's scoring. Through comprehensive
benchmarking with different anomaly detection models at Visa, PersonaAgent is
able to score transactions efficiently (a 1{,}000-transaction CPU batch in
1--2\,ms), lifting the PR-AUC by 0.65--6.58\% and the production business metric
by 8.93--14.24\% relatively in deployment.
\end{abstract}



\maketitle

\section{Introduction}
\label{sec:introduction}

In fiscal year 2025, Visa reported 4.9B credentials and 175M+ merchant locations
across 200+ countries and territories. Visa-branded payment and cash
transactions totaled 329B that year, including those processed by other
networks~\cite{visa2025factsheet}. Transaction anomaly detection at this scale
requires Visa's detection models, which we call detectors, to score each
authorization within a tight latency budget
despite rare anomalies, concept drift, and delayed
\mbox{labels}~\cite{lucas2020credit,pang2021deep,gama2014survey}.
Much of the evidence for an anomaly comes from context: the card's spending
history, merchant behavior, merchant category code (MCC), and location. Payment
systems therefore model entity histories, peer cohorts, and
relations~\cite{yeh2020merchant,xiao2024vecaug,rao2021xfraud}. Joint context also
matters: an MCC that appears unremarkable in aggregate can behave differently
within one location. Transaction foundation models use large-scale
pretraining~\cite{dou2025transactiongpt,yeh2026treasure}, building on earlier
sequential tabular transformers~\cite{padhi2021tabular,zhang2023fata}. These
advances increase what a detector can represent, but do not necessarily shorten
its release cycle.

\begin{figure*}[t]
  \centering
  \includegraphics[width=\textwidth]{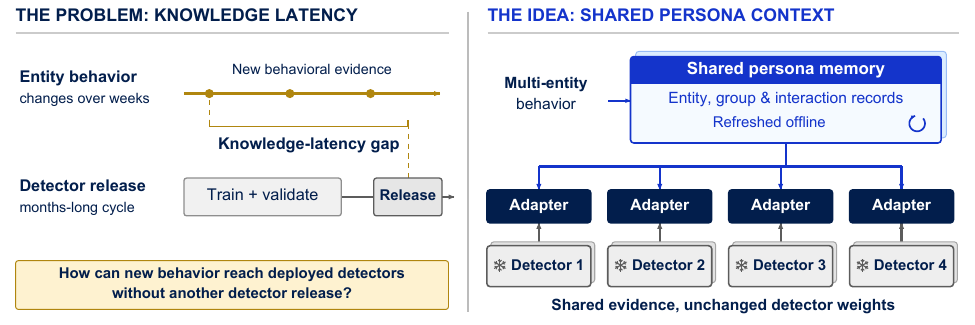}
  \caption{Knowledge-latency motivation and shared persona context. Behavior
  may change before the next detector release. Offline-refreshed entity,
  group, and interaction records supply shared context to detector-specific
  adapters, which combine this context with each detector's own score, while
  the four detectors remain frozen. \mbox{Detectors}~1--4 are Visa's proprietary
  detection models: one production and three internal.}
  \Description{A two-part schematic. The left panel contrasts entity behavior
  changing over weeks with a months-long detector release cycle. The right
  panel shows an offline-refreshed shared persona memory feeding four
  detector-specific adapters, each of which also receives a score from its
  corresponding frozen detector.}
  \label{fig:knowledge-latency}
\vspace{2pt}
\end{figure*}

\enlargethispage{2pt}Industrial detectors such as Visa's are large models. Each release
requires integration and validation~\cite{sculley2015hidden}, so a detector's
update cycle typically lasts at least several months. Entity behavior drifts
much faster; in the setting we study, it can be re-estimated weekly, so the
knowledge encoded in a deployed detector's parameters falls behind. We call this
difference a \textbf{knowledge-latency gap} (Fig.~\ref{fig:knowledge-latency}).
Closing the gap requires behavior evidence that generalizes beyond individual
identifiers and can be refreshed without changing detector weights. Entity
embedding tables~\cite{chen2019behavior}, the common way to learn entity
behavior and interactions, provide limited evidence for rare entities, even less
for entity combinations such as MCC--Location pairs, and none for unseen IDs;
semantic initialization enriches such embeddings~\cite{fan2025enhancing}, but
they remain part of the detector's weights, so updating them still requires a
new release. Feature stores serve refreshed aggregates for individual entities
but leave two questions open: which behavior signals describe an entity, and how
entities with similar behavior should be grouped so that long-tail entities,
whose own evidence is scarce, can borrow from their group. Hand-written rules
and feature-engineering pipelines rely on people to answer these questions
across millions of Merchants, 800+ MCCs, and 230+ Location codes, while
clustering numerical features alone
groups entities by value similarity rather than by what their behavior means.
Large language model (LLM) agents can answer these questions instead: through
\textbf{self-evolution}, they find \textbf{semantic topics}, groups of entities with
similar behavior, and the behavior features that describe them, keeping only
revisions that
improve held-out development performance. Calling an LLM for each transaction
during online scoring
(inference)~\cite{tan2026understanding,lin2026reinforcement,kath2026large},
however, would put text generation on the latency-critical path. These
observations lead to our design: agents learn offline what personas should
capture, and a lightweight adapter injects them into each frozen detector
online.

\begin{figure*}[t]
  \centering
  \includegraphics[width=\textwidth]{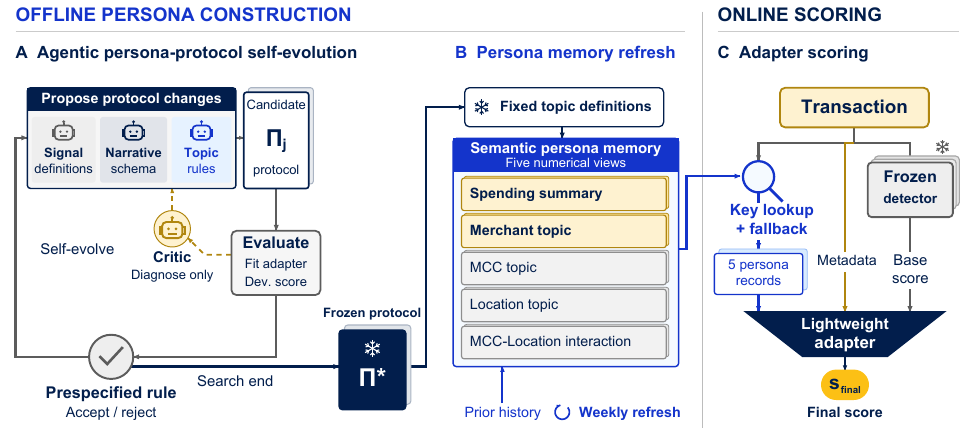}
  \caption{PersonaAgent lifecycle. (A) Three LLM proposal roles and a Protocol
  Critic perform bounded protocol self-evolution. A prespecified rule selects revisions
  using development-set results. (B) Weekly refresh updates numerical persona
  memory under the frozen protocol and fixed topic definitions.
  (C) Exact-key lookup and predefined fallback supply one record for each of
  five views. The adapter combines these with the frozen base score and its
  metadata-adjusted version. LLMs are used only at design time.
  }
  \Description{A three-stage workflow. Offline construction contains a
  role-based protocol-design loop and a weekly refresh for five
  persona views. Online scoring produces one matched or fallback record per
  view, using fixed missing-value indicators when neither is available, and
  combines their numerical feature vector with the frozen
  detector's base score and predefined transaction metadata through a lightweight
  adapter to produce the final score.}
  \label{fig:framework}
  \vspace{6pt}
\end{figure*}

Following this design, we present \textbf{\mbox{PersonaAgent}}, an agentic
framework that stores refreshable multi-entity behavior outside frozen detectors
in \textbf{Semantic Persona Memory}. A \emph{persona} is an auditable behavior
record of an entity, group, or supported interaction at a cutoff time.
PersonaAgent uses LLM agents only at \emph{design time}: an LLM selects salient
signals from
many candidates and describes them in \emph{grounded narratives} that use a
shared behavioral vocabulary. These narratives organize the selected numerical
signals with their temporal and peer-group context, helping agents compare
entity behavior. A narrative schema ties each claim to a declared field, window,
baseline, and support value. An example sentence, with numerical details
omitted, reads: \emph{``Over the declared window, this Merchant's cross-location
activity is above its peer-group baseline and its recent transaction size above
its own prior level.''} The agents search for a persona \emph{protocol}
specifying the signals, narrative schema, and rules for grouping MCCs,
\mbox{Locations}, and Merchants with similar behavior into semantic topics.
Three~LLM roles propose revisions, while a fourth, the Protocol Critic,
diagnoses each candidate. Agents use MCC and Location narratives and their
embeddings to formulate MCC and Location topics, with behavior-feature
clustering forming Merchant topics within the MCC topics. Adapter performance on
a \emph{development set}, held out from all evaluation data, guides revisions to
this hierarchical construction; a prespecified rule selects the protocol to
freeze (Section~\ref{sec:design}). The resulting semantic topics outperform
matched conventional clustering in standalone prediction for three of four group
views (Section~\ref{sec:rq2}). A weekly \emph{refresh} then applies the frozen
protocol to eligible history, updating numerical records under its fixed topic
definitions without an LLM or further revision, while online \emph{scoring}
reads the active validated memory. SR-Fraud~\cite{tan2026srfraud} also uses LLM
agents to adapt detection to changing payment behavior, but calls an LLM at
scoring time to make decisions on transactions. \mbox{PersonaAgent} keeps every
LLM call offline to meet Visa's efficiency and determinism requirements.

\enlargethispage{2pt}The five persona views cover the Card Spending Behavior \mbox{Summary} (\emph{spending
summary}), Merchant topics, MCC topics, Location topics, and supported
MCC-topic--Location-topic interactions. Deterministic lookup by \emph{key} (the
card, a topic, or a topic pair) and fallback supply one record per view. A
lightweight, detector-specific, tree-based adapter then computes the final
score: five encoders, one per view, turn each view's record into an anomaly
prediction, and
a small neural network fuses these predictions with the detector's own score and
its metadata-adjusted version (Section~\ref{sec:inference}). Only the adapter is
fitted; the detector's weights stay unchanged, so new behavioral evidence
reaches each detector without waiting for its next release
(Fig.~\ref{fig:framework}). We evaluate four frozen detectors in a held-out
offline test (Period~1) and a post-launch deployment (Period~2), where live
traffic was scored before outcomes were known. PersonaAgent goes live as a
\emph{release package}: one memory snapshot, the detector-specific adapters that
read it, and release metadata recording their versions. A package is validated
and activated as a whole, so each adapter always reads memory with the schema
and topics it was fitted on. Both periods used the same, unchanged package.
Matched ablations, refit without persona views and evaluated on independent test
data, show the contribution of persona evidence beyond added adapter capacity.
Our contributions are as follows:

\begin{itemize}
  \item We introduce \mbox{PersonaAgent}, an agentic framework for transaction
  anomaly detection that stores multi-entity behavioral evidence in refreshable
  Semantic Persona Memory. LLM agents design the persona protocol offline,
  while memory can be refreshed independently of detector~parameters.
  \item We develop a lightweight retrieval-augmented scoring method that
  enables multiple frozen detectors to use shared persona memory. The method
  combines retrieved behavioral evidence with existing detector scores without
  changing detector parameters or calling an LLM online.
  \item We evaluate \mbox{PersonaAgent} through a held-out offline test and a
  post-launch deployment at Visa. Across four detectors, it achieves an
  8.93--14.24\% relative improvement in the production business metric (BM1)
  during deployment.
\end{itemize}

\section{Related Work}
\label{sec:related}

\paragraph{Transaction representations.}
Transaction foundation models and sequential tabular transformers learn
temporal representations from event histories
\cite{dou2025transactiongpt,yeh2026treasure,skalski2023towards,padhi2021tabular,zhang2023fata}.
Graph models and embedding tables for entities or peer cohorts encode
relationships
\cite{rao2021xfraud,dou2020enhancing,xiao2024vecaug,tang2023gadbench,yeh2020towards},
while payment-network studies incorporate merchant history
\cite{yeh2020merchant}. LLM-based sentence embeddings can also initialize
entity embeddings, enriching representations within the model
\cite{fan2025enhancing}.
Transaction anomaly detection operates under extreme
class imbalance, with confirmed anomalies far rarer than other transactions
\cite{lucas2020credit,pang2021deep}. Related work examines anomaly detection in
financial data and general time series
\cite{he2021mining,paoletti2025mad,liu2025large}, label delay in financial
prediction \cite{sarig2025mind}, and forecasting or imputation under missing or
shifting observations
\cite{yeh2021online,yang2026observations,hu2025bridging,yang2025revisiting,yang2026glocal}.
PersonaAgent pools evidence for low-volume Merchants in behavior-derived
topics, applies minimum-support requirements to group records, and routes
previously unseen Merchants through a predefined parent MCC-topic fallback.
This is related to hierarchical smoothing for high-cardinality categorical
attributes with many distinct values, where a rarely seen value borrows
statistics from its parent group \cite{micci2001preprocessing}.

\paragraph{Retrieval-augmented scoring.}
\enlargethispage{2.3pt}Retrieval-augmented generation grounds a language model's output in documents
retrieved from an external corpus \cite{lewis2020retrieval}, while tabular
retrieval methods use \mbox{similar} rows (nearest neighbors) as additional context
for prediction \cite{qin2021retrieval,gorishniy2024tabr}. BRIGHT separates batch
computation of entity embeddings from real-time prediction within one model
\cite{lu2022bright}. The post-hoc calibration methods in
\cite{guo2017calibration,kull2019beyond} align predictions with observed outcome
rates using only the trained model's output at scoring time. PersonaAgent
retrieves numerical behavioral context from shared, refreshable multi-entity
memory and combines it with each frozen detector's score through a
lightweight, detector-specific adapter. Deterministic key lookup with
predefined fallback serves records built offline, so online scoring requires
no similarity search or generation. The memory and compatible adapters are
validated and released as one versioned package, allowing behavioral evidence to be
updated independently of detector weights.

\paragraph{Agentic design and optimization.}
LLM-based optimizers refine instructions and structured pipelines
\cite{khattab2024dspy,opsahl2024optimizing,yang2024large,zhang2025aflow,shi2025flowxpert}.
CAAFE retains LLM-proposed features only when they improve \mbox{validation}
performance \cite{hollmann2023caafe}, while FeatLLM constructs features offline
so that inference needs no LLM query \cite{han2024featllm}. LLMs can also induce
interpretable topics or clusters from text \cite{pham2024topicgpt,wang2023goal}
or assign retrieval tags~\cite{tang2025llm4tag}. LLM agents forecast time series
\cite{yeh2025empowering}, and a recent survey identifies memory and knowledge
as one capability layer in agentic time-series systems \cite{hulandscape}.
TimeEvo turns a time-series agent's diagnosed failures into new tools and
accepts tool-library revisions only if they pass a paired validation
check~\cite{yang2026timeevo}. PersonaAgent iteratively refines a persona protocol
at design time, with proposal roles suggesting revisions and a Protocol Critic
diagnosing each candidate. This bounded protocol self-evolution determines how
shared, versioned multi-entity memory is built and used by several frozen
detectors. Because LLM evaluators can differ systematically when judging the same evidence under a common rubric
\cite{wang2026understanding}, acceptance follows a prespecified development
rule: a candidate is accepted only when development BM1 improves and the
required checks pass.

\section{Preliminaries and Problem Formulation}
\label{sec:preliminaries}

A transaction $x$ occurs at event time $t(x)$, but its outcome
\mbox{$y(x)\in\{0,1\}$} may not be known until much later. Here, $y(x)=1$ denotes a
confirmed anomaly under the confidential labeling policy. Each snapshot of
Semantic Persona Memory (\mbox{Section}~\ref{sec:memory}) is built from
transactions observed by a cutoff time $T$ and may use an outcome only if it
was available by $T$; we call evidence available by $T$ \emph{cutoff-valid}.
We score $x$ using the release package (memory with compatible adapters) that
was active at its event time $t(x)$, even if the request is processed later.
Its memory was built at a cutoff $T(x)<t(x)$ using only evidence available by
$T(x)$; applying the same rule in offline evaluation therefore prevents scores
from using outcomes learned after the memory cutoff.

A \emph{detector} is a base classifier that scores a transaction from the
transaction's own features. We refer to the four proprietary detectors as
Detectors~1--4. Detector~$m$ is a frozen function $f_m$ with base score
$s_{\mathrm{base}}(x)=f_m(x)\in[0,1]$, where higher means more anomalous; we
treat it as a black box and keep it fixed.

\paragraph{Problem.}
For each detector~$m$, we fit a lightweight adapter that computes a final
anomaly score $s_{\mathrm{final}}(x)$ using three types of information: the
frozen base score $s_{\mathrm{base}}(x)$, the cutoff-valid persona records for
$x$ in the memory built at $T(x)$, and a metadata input derived from a
predefined transaction metadata field (\mbox{Section}~\ref{sec:inference}).
Our goal is to improve anomaly detection performance, as measured by the
metrics below, while keeping $f_m$ unchanged and ensuring that retrieval
and scoring are deterministic and meet the online latency requirement.

\enlargethispage{2.3pt}We report \emph{BM1}, the production business metric at the deployed operating
point, together with PR-AUC and AUC; higher is better for all three. Because
the definition of BM1 and all absolute metric values are confidential, we
report each metric's relative improvement over the frozen detector. For BM1, this is
\mbox{$\Delta\mathrm{BM1}=(\mathrm{BM1}_{\mathrm{final}}-\mathrm{BM1}_{\mathrm{base}})/\mathrm{BM1}_{\mathrm{base}}$};
$\Delta$PR-AUC and $\Delta$AUC are defined likewise, and a relative improvement of $+10\%$
means 1.1 times the base metric value.

\section{Method}
\label{sec:method}

PersonaAgent has three core parts, one for each stage in
Fig.~\ref{fig:framework}: agentic persona-protocol self-evolution, Semantic
Persona Memory, and a tree-based adapter for each detector. At \emph{design
time}
(\mbox{Section}~\ref{sec:design}), LLM roles propose changes to a persona protocol,
and a prespecified rule selects the protocol to freeze. At each weekly
\emph{refresh} (Section~\ref{sec:memory}), a batch job applies the frozen
protocol to cutoff-valid history and~builds a new memory snapshot, which can
be activated only after validation, together with compatible adapters in one
release package. At \emph{scoring time} (Section~\ref{sec:inference}), each
detector's adapter reads five numerical records from the active package and
uses them with the frozen base score and metadata input to compute the final
score. The LLM is used only~at design~time;~the weekly build and online scoring
follow the fixed protocol, and scoring is deterministic. Release controls,
fallback rules, and pseudocode are given in
Appendices~\ref{sec:deployment-controls}, \ref{sec:fallback},
and~\ref{sec:algorithms},~\mbox{respectively.}

\begin{figure*}[t]
  \centering
  \begin{minipage}[t]{0.47\textwidth}
    \centering
    \textbf{(a) MCC topics}\par\vspace{2pt}
    \includegraphics[height=140pt,width=\linewidth,keepaspectratio]{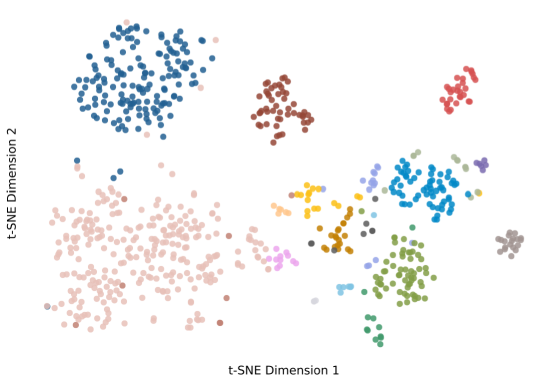}
  \end{minipage}\hfill
  \begin{minipage}[t]{0.47\textwidth}
    \centering
    \textbf{(b) Location topics}\par\vspace{2pt}
    \includegraphics[height=140pt,width=\linewidth,keepaspectratio]{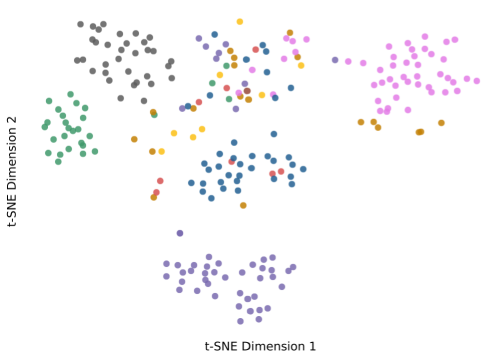}
  \end{minipage}
  \caption{Descriptive visualization of semantic MCC and Location topics on
  two-dimensional t-SNE~\cite{van2008visualizing} maps of narrative embeddings.
  Each equal-size point represents one MCC or Location, and color indicates its
  topic; Locations are anonymized. Section~\ref{sec:rq2} evaluates the topics'
  predictive value.}
  \Description{Two side-by-side scatter plots show MCC and anonymized Location
  narrative embeddings colored by their assigned topics. The visualization
  provides descriptive context rather than a quality or performance test.}
  \label{fig:semantic-topics}
  \vspace{4pt}
\end{figure*}

\subsection{Agentic Persona-Protocol Self-Evolution}
\label{sec:design}

At design time, the agents define how persona memory is built and used during
scoring. These rules form the \emph{persona protocol}, which has five parts:
the behavior signals to compute; the narrative schema, which specifies how
narratives are written; topic construction, which groups MCCs, Locations, and
Merchants with similar measured behavior into \emph{semantic topics}; the
retrieval rules for looking up records and choosing fallbacks; and the release
checks a build must pass. Appendix Table~\ref{tab:protocol-components} gives an
example of each.
To compare measured behavior when defining MCC and Location topics, the agents
use grounded narratives produced under the protocol's narrative schema. For
each narrative, the LLM selects 4--6 numerical behavior signals that it
considers important or indicative of changes in behavior. The schema requires
each claim to name its numerical field, history window, baseline, and support
value (the amount of supporting history). These numerical details constrain
what the narrative can claim while the LLM chooses which behaviors to
emphasize.

\begin{table*}[t]
\caption{Non-sensitive schema for the five persona views. The summary
key identifies the card. Every returned
record also contains its support, valid time interval, source metadata, age,
missing-value status, and fallback level. ``Prior'' means history available
by the construction cutoff; all five views are rebuilt weekly.}
\label{tab:persona-schema}
\centering
\small
\setlength{\tabcolsep}{5.2pt}
\renewcommand{\arraystretch}{1.08}
\begin{tabular}{@{}>{\raggedright\arraybackslash}p{0.18\textwidth}>{\raggedright\arraybackslash}p{0.18\textwidth}>{\raggedright\arraybackslash}p{0.57\textwidth}@{}}
\toprule
Persona view & Key & Main signal families \\
\midrule
Card Spending Behavior Summary & Summary key & Aggregates over twelve months of eligible history; relative and temporal transforms; concentration and diversity; activity distributions (the card's shares of eligible historical activity) over 56 Merchant topics, 19 MCC topics, and nine Location topics \\
Merchant topic & One of 56 topics & Merchants grouped within the parent MCC topic; authorization behavior, channel, cross-location activity, historical transaction-size patterns, outcomes available by cutoff $T$, support, peer-relative deviation, and parent-topic aggregates \\
MCC topic & One of 19 topics & Prior-period levels, trends, recent-to-long-run ratios, robust (outlier-resistant) deviations, ranks, and population-relative behavior \\
Location topic & One of 9 topics & Prior-period geographic behavior with the same temporal and population-relative families as MCC topics \\
MCC--Location interaction & One of up to $19\times9=171$ MCC-topic--Location-topic interactions & Supported interaction or predefined MCC-topic/Location-topic fallback; temporal change and departures from the separate MCC and Location topic patterns \\
\bottomrule
\end{tabular}

\vspace{4pt}
\end{table*}

The protocol is refined and selected through a \emph{protocol search} that
ends before either reported period and uses a separate \emph{development set},
with no data from either period. One LLM performs four offline roles, with a
separate prompt for each role (Fig.~\ref{fig:framework}(A)). The Signal
Definition, Narrative Schema, and Topic Rule roles propose changes to the
behavior signals, narrative schema, and topic construction, respectively. In
round~$j$, these changes are combined into one candidate protocol $\Pi_j$.
For each candidate, the system rebuilds memory and fits a temporary adapter for
\mbox{Detector}~4, after which the \emph{development evaluation} reports development BM1
and check results. Source-consistency checks flag narrative values that do not
match the aggregates they were based on. To guide the next revision, the fourth
role, the Protocol Critic, reviews these flagged values and development BM1.
Its feedback can also lead to changes in the retrieval rules and release
checks.
Under the fixed selection rule, a candidate replaces the current best protocol
only when development BM1 improves and the \mbox{required} checks pass. The
development evaluation, selection rule, and search budget (the maximum number
of candidates) are fixed before the search starts (\mbox{Algorithm}~\ref{alg:design}).
Section~\ref{sec:rq1} tests whether the selected protocol transfers to
Detectors~1--3.

\enlargethispage{2.7pt}The Topic Rule role proposes a hierarchical topic scheme
in which the agents interpret MCC and Location narratives and their embeddings
to select which MCCs and Locations belong to each topic. The embeddings are
numerical vectors that place similar narratives close together
(Fig.~\ref{fig:semantic-topics}). Within each MCC topic, behavior-feature
clustering groups Merchants with similar behavior into Merchant topics at
design time. For example, under one MCC topic, separate Merchant topics
describe predominantly online domestic activity and predominantly online
cross-location activity. Because each candidate's development results guide the
next revision, the MCC and Location topics result from multi-agent
self-evolution rather than a single clustering pass.
Section~\ref{sec:rq2} compares the resulting topic scheme with matched
conventional clustering of numerical features. When the search ends, the
selected protocol $\Pi^\star$, including its topic definitions, is frozen
together with an execution configuration that records the software and
execution settings.

\paragraph{An example topic.} The MCC topic for \textbf{airline and cruise}
travel provides a concrete example, with a narrative that describes behavior
pooled across its member MCCs. With numerical details~omitted here, it reports
average transaction size, card-not-present share, and cross-location share above
their overall baselines, and approval rate below its baseline.
Appendix~\ref{sec:protocol-example} illustrates a \mbox{possible} protocol revision
using this example. Scoring reads the topic's numerical record, not the
narrative, and the weekly build recomputes this record from eligible history
under the same topic definition. The adapter can use this record to score a
purchase at a cruise line the memory has never seen, a \emph{previously unseen}
Merchant. The purchase's MCC still identifies this topic even when the cruise
line has no supported Merchant-topic mapping. Section~\ref{sec:inference}
follows this purchase through scoring.

\subsection{Semantic Persona Memory Refresh}
\label{sec:memory}

With the protocol frozen, weekly builds update the numerical records in
Semantic Persona Memory. Each record stores behavior statistics for one card,
semantic topic, or supported MCC-topic--Location-topic pair, and its key
identifies that card, topic, or pair. A Merchant-topic record, for example,
contains statistics such as the topic's cross-location share. The record is
computed from eligible transactions observed by cutoff $T$, using only outcomes
available by $T$, and its \emph{support} measures how much eligible history was
used. Support decides whether a group has enough history for its own record and
tells the adapter how reliable the record is. Each versioned memory snapshot
contains these records and the maps that assign Merchants, MCCs, and Locations
to their topics.

\begin{figure}[t]
  \centering
  \includegraphics[width=\columnwidth]{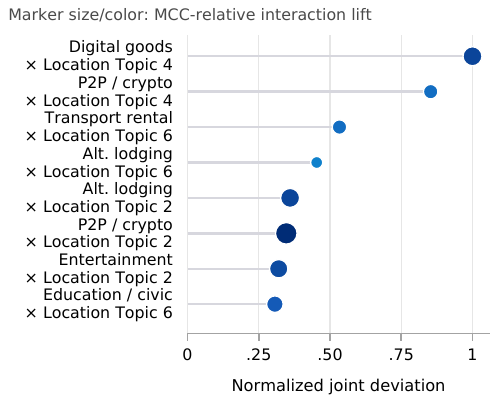}
  \caption{Selected supported MCC-topic--Location-topic interactions from an
  offline snapshot. The horizontal position, normalized within the displayed
  examples, shows the joint deviation (how far a pair's behavior differs from what
  its separate MCC and Location patterns imply); dots grow in area and darken with
  MCC-relative interaction lift, which compares the pair's behavior with that of
  its MCC topic alone. Topic IDs carry no risk
  ordering, and absolute values and Location identities are not disclosed.}
  \Description{A horizontal lollipop chart lists eight anonymized
  MCC-topic--Location-topic interactions. Points farther right have larger
  normalized joint deviation; larger and darker points have higher relative
  interaction lift.
  }
  \label{fig:semantic-interactions}
  \vspace{4pt}
\end{figure}

The memory has five views (\mbox{Table}~\ref{tab:persona-schema}). The spending
summary records twelve months of the card's behavior, including how its
activity is distributed across Merchant, MCC, and Location topics, while the
Merchant, MCC, and Location views summarize behavior in the topics assigned to
the transaction's Merchant, MCC, and Location. The interaction view records how
an MCC-topic--Location-topic pair differs from its separate MCC and Location
patterns.
Topic records combine history across their members, providing statistics for
entities with little or no history of their own. Unlike the per-card spending summaries, group records are
kept for 56 Merchant topics instead of millions of Merchant keys, 19 MCC topics
instead of 800+ MCCs, nine Location topics instead of 230+ Location codes, and
at most $19\times9=171$ topic pairs instead of 184{,}000+ potential
MCC--Location code pairs.

Because an MCC topic can behave differently across Location topics, we keep
a separate record for each supported MCC-topic--Location-topic pair.
Figure~\ref{fig:semantic-interactions} illustrates this variation in one offline
snapshot. Section~\ref{sec:rq2} tests whether the interaction view predicts anomalies on
its own.

\enlargethispage{2.3pt}\textbf{The weekly build.} Each week, the batch job (Fig.~\ref{fig:framework}(B))
uses~eligible historical transactions observed by that week's cutoff $T$ and
outcomes available by $T$. It applies the selected protocol $\Pi^\star$ and its
execution configuration to recompute all five numerical views (Appendix~\ref{sec:persona-fields}). Topics
are settled at design time: the build updates the aggregated feature values
under those topics and maps new \mbox{Merchants} or codes to existing topics by the
fixed topic-assignment rules. Narratives and
embeddings are produced only at design time and are never read during scoring.
The serving memory contains numerical records and their source metadata, and
the weekly build calls no LLM.
Before activation, each build must pass compatibility checks on schema and
topic identity; a failed build leaves the previously validated package active
(Appendix~\ref{sec:deployment-controls}).

\subsection{Online Adapter Scoring}
\label{sec:inference}

All detectors use the same memory, but their scores are on different scales.
Each detector therefore has its own \emph{adapter}, a lightweight model (tree-based encoders and a small neural network) that
combines its base score with information from the retrieved persona
records and
the metadata input to compute a final score (Fig.~\ref{fig:framework}(C)); the
detector itself is never changed. Scoring retrieves one persona record per view,
forms the adapter's inputs, and runs the adapter.

\textbf{Retrieving persona records.} The card's own key selects its spending
summary, the topic-assignment maps
assign the transaction's Merchant, MCC, and \mbox{Location} to their topics, and the
pair of MCC and Location topics forms the interaction key. Exact-key lookup then
returns one numerical record per view, so the number of retrievals is fixed and
scoring involves no nearest-neighbor search or topic construction. When a
Merchant has no supported topic mapping or a group has too little history for
its own record, a predefined fallback record from a
larger group fills the view (\mbox{Appendix}~\ref{sec:fallback}).

\textbf{Adapter inputs.} From each record, the adapter takes a fixed set of
numerical fields, including behavior values and reliability indicators such as
support and fallback level. Together with three activity shares from the
spending summary (the card's shares of eligible historical activity in the
transaction's Merchant, MCC, and Location topics), these fields form the
\emph{persona feature vector}, whose length is fixed regardless of how many
cards and Merchants the memory covers
(\mbox{Appendix}~\ref{sec:retrieval-contract}). The system computes the
\emph{metadata input} by comparing a predefined transaction metadata field with
a memory-derived Merchant-topic reference.

\textbf{The example purchase.} For the cruise purchase of Section~\ref{sec:design}, the MCC maps to the
airline-and-cruise topic and the Location to a Location topic, which together
give its interaction key; the cruise line itself has no supported mapping to a
Merchant topic. The MCC view therefore reads the airline-and-cruise topic's
record, and the Merchant view uses the parent MCC-topic fallback record for the
same topic. In the persona feature vector, the MCC share is the card's share of
eligible historical activity in the airline-and-cruise topic, or 0 if the card
has none there; the Merchant share is marked missing, not replaced by the MCC
share. Section~\ref{sec:cost} shows that PersonaAgent's gain holds on transactions served
by MCC-topic fallback, such as this one.

\begin{table}[t]
\caption{Design, fitting, and evaluation stages. Detectors remain frozen
throughout. Neither reported period is used for adapter or protocol selection.}
\label{tab:evaluation-stages}
\centering
\footnotesize
\setlength{\tabcolsep}{3pt}
\renewcommand{\arraystretch}{1.05}
\begin{tabular}{@{}>{\raggedright\arraybackslash}p{0.25\columnwidth}>{\raggedright\arraybackslash}p{0.70\columnwidth}@{}}
\toprule
Stage & What is fitted or held fixed \\
\midrule
Protocol design & Each candidate rebuilds memory and fits a temporary Detector~4
adapter for development evaluation. \\
\addlinespace[2pt]
Final fitting & The selected protocol governs persona construction; four
final adapters are fitted and selected. One compatible package is frozen
before Period~1. \\
\addlinespace[2pt]
Period~1 & The frozen package is evaluated on held-out offline transactions. \\
\addlinespace[2pt]
Period~2 & The same package scores live traffic; outcomes are joined later.
No snapshot or adapter is replaced. \\
\bottomrule
\end{tabular}
  \vspace{4pt}
\end{table}

\textbf{Adapter model.} Five tree-based encoders, one per persona view, handle
the different fields and missing values across views.
Each encoder takes the features of one view and produces an anomaly prediction
between 0 and 1; the spending-summary encoder also uses the three activity shares.
The metadata input enters the adapter only through the
metadata-adjusted base score $\widetilde s_{\mathrm{base}}(x)$.
The fusion network $\operatorname{Fuse}$, a small multilayer perceptron (MLP)
with hundreds of parameters, models interactions among the five view predictions, the base score, and the
metadata-adjusted base score to compute the final score
(\mbox{Appendix}~\ref{sec:deployment-controls}):
\begin{equation}
\begin{aligned}
s_{\mathrm{final}}(x)=\operatorname{Fuse}\big(
&\!\operatorname{logit}u_{\mathrm{cb}}(x),\operatorname{logit}u_{\mathrm{mer}}(x),\\[-2pt]
&\!\operatorname{logit}u_{\mathrm{mcc}}(x),\operatorname{logit}u_{\mathrm{loc}}(x),\\[-2pt]
&\!\operatorname{logit}u_{\mathrm{int}}(x),s_{\mathrm{base}}(x),\widetilde s_{\mathrm{base}}(x)\big),
\end{aligned}
\label{eq:fusion-input}
\end{equation}
where each $u$ is the prediction from one view, with subscripts $\mathrm{cb}$ for
the spending summary and $\mathrm{int}$ for the interaction view.
Final adapters are fitted offline before Period~1; a weekly
build is released with them only if it passes the compatibility checks
(\mbox{Section}~\ref{sec:memory}).
\mbox{Starting} from precomputed base scores, the online path scores a
1{,}000-transaction CPU batch in 1--2\,ms, with no LLM call or nearest-neighbor
search (Section~\ref{sec:cost}).

\section{Evaluation}
\label{sec:evaluation}

Following the setup in Section~\ref{sec:setup}, we address four research
questions (RQs) in \mbox{Sections}~\ref{sec:rq1}--\ref{sec:cost}. RQ1 asks
whether PersonaAgent improves all four frozen detectors, offline and in
deployment. RQ2 asks how informative the persona views and semantic topics are,
RQ3 which alerts the final score would change in deployment, and RQ4 how
efficient and robust PersonaAgent is in operation.

\subsection{Evaluation Setup}
\label{sec:setup}

To evaluate the contribution of persona memory to detectors already in use,
we compare each detector's final score with its own frozen base score. The
comparison measures score quality, not authorization decisions. The BM1
operating-point rule was fixed before Period~1 and is applied unchanged to the
base and final scores in both periods.
Table~\ref{tab:evaluation-stages} separates protocol design and adapter
fitting from the two evaluation periods. Period~1 is a held-out offline test;
Period~2 scored live traffic before any outcome was known, and outcomes were
joined later. The memory snapshot and four adapters frozen before Period~1 stay
unchanged through Period~2, which tests whether gains over the frozen
detectors persist as the memory ages without weekly refresh.

\subsection{RQ1: Does PersonaAgent Improve Four Frozen Detectors?}
\label{sec:rq1}

The release package improves every detector on every metric in both periods.
Table~\ref{tab:business-results} reports relative improvements over each
frozen detector: all 24 detector--period--metric comparisons show improvements, and
every 95\% confidence interval lies above zero. These intervals are computed
with a paired bootstrap over transactions (Appendix~\ref{sec:uncertainty}).

\begin{table}[t]
\caption{Relative improvement (\%) over each frozen detector, with 95\% confidence intervals.
Period~1: held-out offline test; \mbox{Period}~2: deployment.
$^\dagger$Design-time protocol selection used development-set feedback from
Detector~4; neither reported period was used for selection.}
\label{tab:business-results}
\centering
\small
\setlength{\tabcolsep}{3pt}
\renewcommand{\arraystretch}{1.09}
\newcommand*{\liftci}[2]{{\scriptsize$[#1,\,#2]$}}
\begin{tabular}{@{}cl*{3}{r@{\hspace{2pt}}l}@{}}
\toprule
Detector & Period & \multicolumn{2}{c}{$\Delta$BM1} & \multicolumn{2}{c}{$\Delta$PR-AUC}
  & \multicolumn{2}{c@{}}{$\Delta$AUC} \\
\midrule
1 & Period~1 & $+15.03$ & \liftci{11.91}{19.22} & $+1.40$ & \liftci{1.28}{1.52} & $+0.27$ & \liftci{0.25}{0.29} \\
    & Period~2 &  $+8.93$ & \liftci{7.73}{10.23} & $+1.30$ & \liftci{1.19}{1.42} & $+0.27$ & \liftci{0.25}{0.29} \\
\addlinespace[2pt]
2 & Period~1 & $+20.08$ & \liftci{16.99}{22.69} & $+4.91$ & \liftci{4.72}{5.09} & $+0.73$ & \liftci{0.70}{0.75} \\
    & Period~2 & $+14.24$ & \liftci{12.43}{15.98} & $+3.53$ & \liftci{3.31}{3.75} & $+0.66$ & \liftci{0.64}{0.69} \\
\addlinespace[2pt]
3 & Period~1 &  $+4.99$ & \liftci{4.13}{5.91} & $+0.69$ & \liftci{0.60}{0.77} & $+0.28$ & \liftci{0.26}{0.30} \\
    & Period~2 & $+11.15$ & \liftci{8.65}{13.60} & $+0.65$ & \liftci{0.55}{0.74} & $+0.23$ & \liftci{0.22}{0.24} \\
\addlinespace[2pt]
4\rlap{$^\dagger$} & Period~1 & $+11.76$ & \liftci{9.88}{13.24} & $+5.58$ & \liftci{5.38}{5.77} & $+0.40$ & \liftci{0.38}{0.42} \\
    & Period~2 & $+10.86$ & \liftci{9.31}{12.65} & $+6.58$ & \liftci{6.30}{6.84} & $+0.49$ & \liftci{0.47}{0.51} \\
\bottomrule
\end{tabular}
  \vspace{4pt}
\end{table}

All 24 comparisons also show improvements when the fusion network is refitted on the
same data with five random seeds. Protocol search used only
\mbox{Detector}~4 feedback, so Detectors~1--3 test transfer of the selected
protocol. Detector~2 achieves the largest gain on the selection metric BM1 in
both periods, showing that the protocol's benefit extends beyond the detector
used to select it.

In deployment, BM1 improves by a relative 8.93--14.24\%, while the
threshold-free PR-AUC and AUC improve by 0.65--6.58\% and 0.23--0.66\%,
respectively. The gains therefore extend from the deployed operating point to
both ranking metrics, with smaller relative improvements in AUC.

From Period~1 to Period~2, relative improvements fall in eight of the twelve
detector--metric combinations, are unchanged in one, and rise in three. The
relative improvement in BM1 falls for Detectors~1, 2, and~4 but more than doubles for
\mbox{Detector}~3 (Appendix~\ref{sec:period-analysis}). The unchanged release
package nonetheless improves all four frozen detectors in
\mbox{Period}~2.

\subsection{RQ2: How Informative Are Persona Views and Semantic Topics?}
\label{sec:rq2}

\paragraph{Contribution of persona views.}
RQ1 shows that the release \mbox{package} improves every detector; RQ2 first uses
matched \mbox{ablations} to assess what the five persona views add to the adapter
beyond its base-score inputs. Each variant refits the fusion network on the
same training rows, changing only its inputs, and scores the transactions of
both periods offline (\mbox{Appendix}~\ref{sec:matched-controls}). With the frozen base score as its only
input, the refit adapter leaves PR-AUC and AUC unchanged (0.00\% relative improvement) for every
detector in both periods. A second variant adds the metadata-adjusted base score
but none of the five persona views; it slightly reduces PR-AUC for
\mbox{Detectors}~1--3 and gives a $+1.17\%$ relative improvement for
\mbox{Detector}~4. Adding the five persona
views to this second variant increases the relative improvements in PR-AUC and AUC by 3.20 and
0.44 percentage points on average, and PR-AUC, AUC, and BM1 all rise in each of
the eight detector--period combinations.

\begin{figure}[t]
  \centering
  \includegraphics[width=\columnwidth]{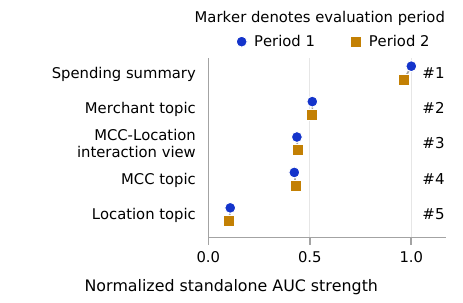}
  \caption{Normalized standalone AUC strength of single-view predictors trained
  before Period~1 and frozen across both periods. AUC above chance is divided
  by that of the spending summary in Period~1. Both periods share this scale:
  0 denotes chance, and 1 matches the reference, not perfect prediction.}
  \Description{The Card Spending Behavior Summary ranks first in both periods, followed by
  Merchant topic, MCC--Location interaction view, MCC topic, and Location topic.}
  \label{fig:persona-standalone-auc}
  \vspace{-8pt}
\end{figure}

\paragraph{Standalone view prediction.}
We next assess the information in each view alone. Each single-view
predictor scores transactions from one persona view, without the base score or
the fusion network, on a pre-release memory snapshot built under the released
package's protocol and topic definitions. Both figures report
\emph{\mbox{normalized} strength}: a predictor's AUC above chance,
$\mathrm{AUC}-0.5$, divided by that of a reference predictor. Here the reference
is the spending summary in Period~1; the topic comparison below uses a
different reference, so values should be compared within each figure.
All five views predict anomalies above chance in both periods, and their
ranking holds as the memory ages: the spending summary first, followed by
Merchant topics (Fig.~\ref{fig:persona-standalone-auc}). The top-ranked spending
summary itself draws on the topics: it records the card's shares of eligible
historical activity across Merchant, MCC, and Location topics, such as its
share in the MCC topic for airline and cruise travel
(Appendix~\ref{sec:persona-fields}). \enlargethispage{3pt}The interaction predictor ranks slightly
above the MCC predictor and well above the Location predictor in both periods,
so joint MCC-topic--Location-topic context, kept for supported pairs
(Section~\ref{sec:memory}), predicts better than either topic alone.

\begin{figure}[t]
  \centering
  \includegraphics[width=\columnwidth]{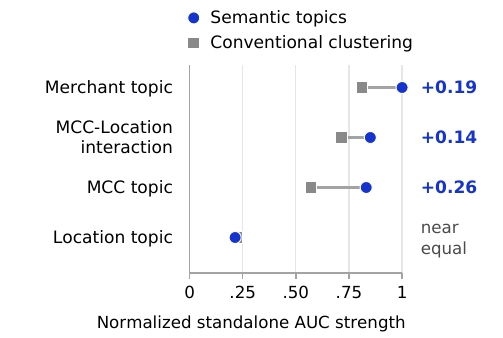}
  \caption{Semantic topics versus matched conventional clustering in
  Period~1, by group view. AUC above chance is normalized by that of the semantic Merchant-topic
  predictor, the largest plotted value: 0 denotes chance and 1 matches this
  reference. Labels give the differences in normalized strength; Location
  scores are nearly equal.}
  \Description{A horizontal dumbbell chart compares semantic topics with
  conventional clustering in Period~1 for Merchant, MCC, Location, and
  MCC--Location interaction views derived from their respective groupings. Semantic topics have higher normalized scores in three views; Location scores are
  nearly equal.}
  \label{fig:topic-ablation}
  \vspace{4pt}
\end{figure}

\paragraph{Semantic topics versus conventional clustering.}
We finally assess the hierarchical topic grouping as a whole by comparing the
topics selected at design time with conventional clustering for the Merchant,
MCC, Location, and interaction views. Conventional clustering forms groups
directly from numerical features without narrative embeddings or
agent-formulated topic definitions. Both groupings use the same held-out
Period~1 transactions, numbers of groups, single-view predictor, and evaluation
procedure, and each derives its interaction view from its own MCC and Location
groups. The selected scheme follows the hierarchical construction of
Section~\ref{sec:design}: agent-formulated MCC and Location topics, mapped in
Fig.~\ref{fig:semantic-topics}, with Merchant topics formed by behavior-feature
clustering within the MCC topics. In standalone prediction, the selected topic
scheme outperforms matched conventional clustering for three of the four group
views (Fig.~\ref{fig:topic-ablation}). Here normalized strength uses the semantic
Merchant-topic predictor's AUC above chance as its reference. On this scale,
semantic topics lead by $+0.26$ on MCC (about a quarter
of the reference's AUC above chance), $+0.19$ on Merchant, and $+0.14$ on the
interaction view; Location scores are nearly equal.

\subsection{RQ3: Which Alerts Would the Final Score Change in Deployment?}
\label{sec:deployment}

RQ3 examines individual alert changes in addition to aggregate metrics. The
base and final scores were recorded live in Period~2, before outcomes were
known. Once outcomes became available, we reconstructed alerts from those
recorded scores using the same BM1 operating-point rule fixed before
Period~1, with a separate analysis threshold determined for each score.
Compared with the base score's alerts, the final score can recover a missed
anomaly (a \emph{recovery}), withdraw a false alert, introduce a false alert,
or lose a detected anomaly (a \emph{lost detection}). The relative improvements in BM1 in
Table~\ref{tab:business-results}, \mbox{measured} at this same operating point, show
the combined effect of these four changes.

Recovered anomalies are largely detector-specific: of the distinct anomalies
recovered for at least one detector in Period~2, 71.7\% are recovered for
exactly one detector and 10.6\% for three or more. To compare correction
patterns across periods, we use a detector's \emph{correction fingerprint},
which contrasts its recovered anomalies with its lost detections on
characteristics that include the five persona-view scores. For every
detector, the Period~1 and Period~2 fingerprints have a cosine similarity of
at least 0.93, \mbox{indicating} similar patterns of relative differences in these~characteristics across periods (Appendix~\ref{sec:correction-fingerprint}).
Table~\ref{tab:deployment-cases} illustrates these corrections with two
recovered anomalies and two withdrawn false alerts from Period~2, with
persona-view scores given as~percentiles (pct).

\begin{table}[t]
\caption{Four sanitized Period~2 cases at the retrospective BM1 operating
point. FN: an anomaly the frozen detector missed; FP: a false alert it raised.
\emph{pct}$n$ gives the percentile of a view's prediction score before fusion,
among that detector's Period~2 transactions whose alert status changed (all
four change types).}
\label{tab:deployment-cases}
\centering
\footnotesize
\setlength{\tabcolsep}{2.2pt}
\renewcommand{\arraystretch}{1.02}
\begin{tabular}{@{}
  >{\raggedright\arraybackslash}p{0.095\columnwidth}
  >{\raggedright\arraybackslash}p{0.4125\columnwidth}
  >{\raggedright\arraybackslash}p{0.4125\columnwidth}@{}}
\toprule
Error & Transaction context & Persona evidence and alert change \\
\midrule
\textbf{FN}
& A money transfer with low group-view scores (Merchant~pct15, MCC pct3,
interaction pct9) and a \$100--200 amount well below its Merchant topic's
usual level.
& The spending-summary score was at pct99. PersonaAgent recovered this confirmed
anomaly. \\
\addlinespace[2pt]
\textbf{FN}
& A charge at a direct-marketing digital-goods merchant, with \mbox{a spending-summary score} at pct49 and a \$100--200 amount well above its
Merchant topic's usual~level.
& The Merchant-view score was at pct90. PersonaAgent recovered this confirmed
anomaly. \\
\addlinespace[2pt]
\textbf{FP}
& A charge sharing the Merchant-topic record above, with view scores at pct90
(Merchant), pct55 (MCC), and pct44 (interaction), and an amount under \$10.
& The spending-summary score was at pct3. PersonaAgent withdrew this false
alert. \\
\addlinespace[2pt]
\textbf{FP}
& A transaction with Merchant- and interaction-view scores at pct83 and pct88
and an amount above \$500, far above its Merchant topic's usual level.
& The spending-summary score was at pct7. PersonaAgent withdrew this false
alert. \\
\bottomrule
\end{tabular}
  \vspace{-10pt}
\end{table}

\enlargethispage{3pt}In three of the four cases (rows one, three, and four), the alert changes are
consistent with the card's spending-summary scores despite contrasting
group-view scores. Rows two and three share a Merchant-topic record and a Merchant-view
score at pct90, but have different spending-summary scores and opposite alert
changes: row two, with a mid-range spending-summary score (pct49), is a
recovered anomaly, whereas row three, with a low score (pct3), is a withdrawn
false alert. The anomaly in row one is recovered despite low group-view scores,
while the false alert in row four is withdrawn despite high group-view scores. Their
spending-summary scores are pct99 and pct7, respectively; the transaction
amount is below the Merchant topic's usual level in row one and far above it
in row four.

\subsection{RQ4: How Efficient and Robust Is PersonaAgent in Operation?}
\label{sec:cost}

We measure serving cost and performance under sparse history, then describe
how scoring and refresh handle failures.

\paragraph{Serving cost.}
On CPU, the online path scores a batch of 1{,}000 transactions in 1--2\,ms,
measured from receiving the transactions and their frozen base scores to
returning final scores; this covers lookup, feature-vector construction, the
five encoders, and
the fusion of Eq.~\ref{eq:fusion-input}. Adding another detector reuses the
shared memory build and requires its own adapter fit; fitting one adapter took
approximately 20--48 minutes.

\paragraph{Sparse history.}
On transactions served by MCC-topic fallback (\mbox{Appendix}~\ref{sec:fallback}),
such as those at previously unseen Merchants, the mean relative improvement in PR-AUC across the four detectors is 3.30\% and 3.64\% in \mbox{Periods}~1 and~2. All 24 detector--period--metric comparisons show improvements with
confidence intervals above zero when the comparison in \mbox{Table}~\ref{tab:business-results} is
recomputed on this subset. These results address the long-tail case raised in
Section~\ref{sec:introduction}.

\paragraph{Failure handling.}
Failures in PersonaAgent's path do not stop scoring. A serving failure, such
as an unavailable lookup, returns the frozen base score; a failed refresh
validation leaves the previous package active, so scoring continues with
the existing memory. Figure~\ref{fig:deployment} shows release control and
request serving, and \mbox{Appendix} Table~\ref{tab:failure-semantics} gives the
corresponding controls.

\section{Discussion}
\label{sec:discussion}

\paragraph{Design principles.}
Compatibility with an existing adapter requires stable topic meanings as well
as stable input dimensions. The spending summary, for example, stores a card's
share of past activity in each topic. If a topic ID were reused for different
behavior, the adapter would still accept its inputs but misread them: a card's
share of the redefined topic would count as its share of the original one.
Topic identity therefore belongs in release checks alongside schema and
fallback. The validated release package is the unit of activation, maintenance,
and rollback, allowing entity knowledge to be updated independently of each
detector's release cycle. This separates scheduled knowledge construction from
serving activation: a weekly build produces a candidate memory, but activation
requires a validated package rather than a new detector release.

\paragraph{Scope and future directions.}
The record-age comparison shows that the relative improvement in BM1 is
lower for every detector
when MCC-topic, Location-topic, and interaction records come from the
preceding weekly build (Appendix~\ref{sec:record-age}). This sensitivity to
record age motivates a direct comparison of weekly refresh with a fixed
release package. Versioned releases allow the refresh frequency to be varied,
so the performance benefit of fresher records can be evaluated against rebuild
and validation cost.
Location-derived signals may correlate with sensitive attributes, so
subgroup-impact assessment should precede policy-level use.

\section{Conclusion}

PersonaAgent addresses the knowledge-latency gap between weekly re-estimation
of entity behavior and months-long \mbox{detector}~release cycles. It keeps behavioral
context in a shared, refreshable persona memory outside the detectors,
organized by semantic topics that pool evidence for rare and cold-start
entities. At design time, the protocol for building this memory evolves through
a bounded agentic search: LLM roles propose revisions, and a prespecified rule
selects the protocol to freeze. Online, deterministic retrieval and
detector-specific adapters link the memory to frozen detectors without LLM
calls. One release package, unchanged across both periods, improved every reported
metric for four detectors offline and in deployment, where the production
business metric (BM1) rose by a relative 8.93--14.24\%. New behavioral evidence
can reach several deployed detectors through one validated release package,
without waiting for any detector's next release.

\begin{figure}[t]
  \centering
  \includegraphics[width=\columnwidth]{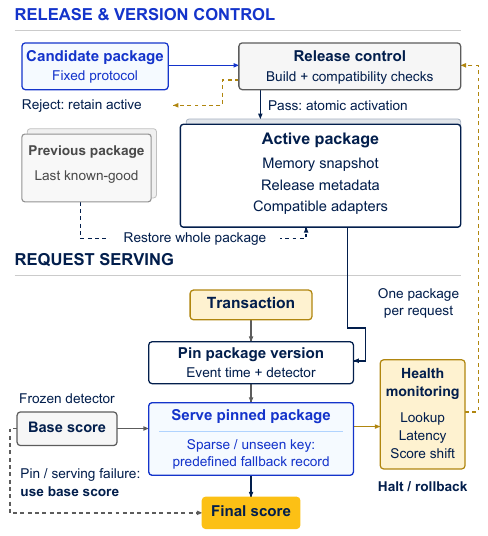}
  \caption{Release control and request serving. Memory, release metadata, and
  compatible adapters are activated atomically as one package only after build
  and compatibility checks pass. Each scoring request selects one package
  version by event time and detector; health monitoring can halt a rollout or
  restore the complete previous package. If package selection or serving
  fails, the frozen base score is returned.
  }
  \Description{A two-part system diagram. Release control validates a candidate
  package, activates it as a unit on success, and retains the active package on
  rejection. A previous complete package is retained for rollback. Each request
  pins a compatible package using event time and detector identity before serving.
  Sparse or unseen keys use predefined fallback records and continue scoring;
  pinning or serving failures instead return the frozen base score.
  Lookup, latency, and score-shift monitoring feeds back to release control.}
  \label{fig:deployment}
  \vspace{4pt}
\end{figure}

\setlength{\bibsep}{0pt plus 0.15pt}
\bibliographystyle{ACM-Reference-Format}
\bibliography{anomaly_refs}

\appendix
\section{Privacy and Compliance Statement}
\label{sec:governance}

The evaluation uses access-controlled derived records and internal lookup keys. The paper
reports no primary account number, transaction sequence identifier, raw
merchant name, Location identity, exact transaction/cohort count, detector
identity, operating threshold, or absolute performance value. Merchant-, MCC-,
and Location-topic identifiers have no risk ordering. Card Spending Behavior
Summaries describe payment behavior rather than demographic or psychological
attributes.
The proprietary transaction data, model artifacts, prompts, code, and production
configuration remain confidential.
This paper documents the data interfaces, lifecycle stages, non-sensitive
signal families, algorithms, and relative results without exposing proprietary
data or implementation details.

\section{Deployment Controls and Failure Handling}
\label{sec:deployment-controls}

Operational controls must prevent an invalid refresh from reaching online
scoring unnoticed. Figure~\ref{fig:deployment} shows the boundary between
offline validation and online retrieval from an activated release package;
Algorithms~\ref{alg:refresh}--\ref{alg:serve} give the activation and serving
steps, and Table~\ref{tab:failure-semantics} summarizes the corresponding failure
responses. We check three aspects separately. Build validation
checks whether evidence was available by the cutoff, whether the schema is
complete, and whether support, topic identities, and freshness meet the
predefined requirements. Compatibility
validation confirms that the memory snapshot provides the topic identities,
fields, dimensions, ranges, and fallback behavior expected by the adapter.
Online health monitoring tracks successful exact-key lookups, fallback and missing rates,
snapshot age, lookup and adapter latency, and changes in score distributions.
A package is activated only after it passes the build and compatibility checks;
online health monitoring can halt a rollout or trigger rollback.

Fallback records are explicit protocol outputs.
Their fallback level, support, age, and missing-value status are passed to the
adapter. If any step of the memory path fails, the system returns the frozen
detector's score. After validation,
activation switches the complete compatible release package as one unit. Rollback
restores the previous release package. The active-package registry $\mathcal A$ records when each package
was active and supports later reconstruction using only the information
available at each event time. This reconstruction serves audits; Period~2 scores
were produced live, before any outcome was known.

\begin{table}[htbp]
\caption{Failure and fallback conditions in PersonaAgent's path and the safe
action for each.}
\label{tab:failure-semantics}
\centering
\footnotesize
\setlength{\tabcolsep}{3.2pt}
\renewcommand{\arraystretch}{1.08}
\begin{tabular}{@{}>{\raggedright\arraybackslash}p{0.27\columnwidth}>{\raggedright\arraybackslash}p{0.31\columnwidth}>{\raggedright\arraybackslash}p{0.32\columnwidth}@{}}
\toprule
Condition & Observable evidence & Safe action \\
\midrule
Incomplete or stale build & completeness/\allowbreak freshness checks & keep the last
active release package \\
\addlinespace[3pt]
Sparse or unseen key & support and fallback-level indicators & use predefined fallback;
preserve its level \\
\addlinespace[3pt]
Memory lookup unavailable & lookup health and timeout & return the frozen
base score \\
\addlinespace[3pt]
Unexpected score shift & score-change and no-regression checks & halt
activation or restore the prior release package \\
\bottomrule
\end{tabular}
\vspace{-11pt}
\end{table}

\paragraph{Adapter architecture and fitting.} Each persona view contributes one encoder
output; the original and metadata-adjusted~base~scores supply the other two
inputs. We use tree-based encoders and an~MLP fusion network because trees
remain strong on tabular data~\cite{chen2016xgboost,grinsztajn2022tree} and
tree/MLP hybrids are effective~\cite{yan2024team}. The deterministic fusion
network contains hundreds
of parameters, excluding the tree-based encoders. In Eq.~\ref{eq:fusion-input}, each
encoder output is first clipped slightly away from 0 and 1 so that its logit is
finite. Final adapters are fitted and selected using eligible outcomes and
cutoff-valid persona feature vectors. The fusion network is trained on
out-of-fold encoder predictions: we
split the training rows into five parts and predict each part with encoders
fitted on the other four.

\section{Persona Records, Retrieval, and Fallback}
\label{sec:persona-fields}

Table~\ref{tab:record-contract} summarizes persona-record fields and retrieval
metadata across the five persona views.

\begin{table}[t]
\caption{Non-sensitive fields associated with persona records and retrieval
results. Exact field names and internal limits are implementation details.}
\label{tab:record-contract}
\centering
\footnotesize
\setlength{\tabcolsep}{3pt}
\begin{tabular}{@{}>{\raggedright\arraybackslash}p{0.24\columnwidth}>{\raggedright\arraybackslash}p{0.66\columnwidth}@{}}
\toprule
Field group & Information \\
\midrule
Identity and time & Persona view; summary, topic, or interaction key;
valid interval; construction cutoff; memory-snapshot and topic-scheme identifiers \\
\addlinespace[3pt]
Evidence & Eligible aggregate values; relative ratios; temporal change;
population/topic-relative ranks and robust deviations \\
\addlinespace[3pt]
Reliability & Support, missing-value status, memory age, fallback eligibility,
actual fallback level, and validation results \\
\addlinespace[3pt]
Topic construction & Design-time narrative-schema and
embedding-configuration IDs, topic identity, parent identity, and
identity-consistency status \\
\addlinespace[3pt]
Source tracking & Protocol ID, execution configuration, release metadata, input-data lineage, build identifier, and activation decision \\
\bottomrule
\end{tabular}
  \vspace{-5pt}
\end{table}

\paragraph{Memory construction.}
At cutoff $T$, the weekly build of \mbox{Section}~\ref{sec:memory} computes the
memory
\begin{equation}
K_T=\mathcal C_{\Pi^\star,E^\star}\!
\left(\mathcal X_T,\mathcal Y_T\right),
\label{eq:memory-build}
\end{equation}
that is, the build $\mathcal C$ applies the frozen protocol $\Pi^\star$, with
its fixed topic definitions, under the execution configuration $E^\star$ to the
eligible historical transactions $\mathcal X_T$ observed by $T$ and their
outcomes $\mathcal Y_T$ available by $T$. Algorithm~\ref{alg:refresh} writes
$D_T$ for this evidence and $K_T^{\mathrm{cand}}$ for the snapshot before
validation.
The five views are indexed by $\mathcal L=\{\mathrm{cb,mer,mcc,loc,int}\}$, and
the topic scheme has three families,
$\mathcal L_{\mathrm{topic}}=\{\mathrm{mer,mcc,loc}\}$.
The spending summary covers the card's own history over these topics; the
interaction
view uses supported MCC-topic--Location-topic pairs. All five views
and their topic-assignment maps share this cutoff.

\paragraph{Retrieval keys.}
\label{sec:retrieval-contract}
Online retrieval uses exact-key lookup over a versioned snapshot rather than
nearest-neighbor search. Using a snapshot built at cutoff $T$, the system
resolves five logical keys for transaction $x$, one per view
(Fig.~\ref{fig:framework}(C)):
\begin{equation}
\begin{aligned}
\kappa_{T}(x)=\big(&k_{\mathrm{cb},T}(x),k_{\mathrm{mer},T}(x),
k_{\mathrm{mcc},T}(x),\\[-2pt]
&k_{\mathrm{loc},T}(x),k_{\mathrm{int},T}(x)\big).
\end{aligned}
\label{eq:keys}
\end{equation}
The spending summary is keyed by the card itself; Merchant, MCC, and Location
fields resolve through the topic-assignment maps, and
$k_{\mathrm{int},T}(x)=\big(k_{\mathrm{mcc},T}(x),k_{\mathrm{loc},T}(x)\big)$
is the resolved MCC-topic--Location-topic pair.

\paragraph{Lookup and feature vector.}
\begin{align}
 r_\ell(x)
&=\operatorname{LookupOrFallback}\bigl(
K_{T(x)},\ell,\notag\\
&\qquad\kappa_{T(x)}(x);\beta_{\ell,T(x)}\bigr),
\label{eq:lookup}\\
\phi(x)
&=\operatorname{Project}\!\left(
\{r_\ell(x)\}_{\ell\in\mathcal L},\kappa_{T(x)}(x)\right),
\label{eq:retrieve}
\end{align} where $\operatorname{LookupOrFallback}$ follows the predefined
fallback map
$\beta_{\ell,T(x)}$, and $\operatorname{Project}$ returns a fixed-length feature
vector:
each view's record values and reliability indicators, plus the three activity
shares the resolved topic IDs select from the spending summary. Its length never
changes as the population grows.

\paragraph{Fallback rules.}
\label{sec:fallback}
A group must meet the protocol's minimum support requirement to have its own
record; otherwise, $\operatorname{LookupOrFallback}$ follows
$\beta_{\ell,T(x)}$, which the snapshot stores, to a record for a larger group.
A
Merchant without a supported topic mapping uses its parent MCC-topic fallback
record for the Merchant view, while the MCC view is still read separately with
the MCC-topic key; this fills the Merchant view without assigning the Merchant
to a Merchant topic. The metadata input for such a Merchant uses the parent
MCC-topic reference. If an MCC-topic--Location-topic pair has too little
history, the interaction view uses a predefined record for one of the two
topics (a marginal record), such as the MCC topic, or a fixed missing-value
record if no such record is available. Each view keeps its fixed schema, and
the fallback level (exact, parent, marginal, or missing) tells the adapter what
evidence was available. Sparse history is therefore handled by fallback; only a
serving failure returns the frozen base score. \mbox{Algorithm}~\ref{alg:serve}
gives the scoring pseudocode, including how
such failures are handled. Section~\ref{sec:cost} shows that the gain holds on
transactions served by MCC-topic fallback.

\paragraph{Card spending behavior for the current transaction.}
The retrieved spending summary stores empirical shares of eligible historical
activity across the Merchant, MCC, and Location topic families, written
$p_{\mathrm{cb},T(x)}^\ell$. For an unresolved key $\bot$ in topic family
$\ell\in\mathcal L_{\mathrm{topic}}$, \mbox{define}
$p_{\mathrm{cb},T(x)}^\ell[\bot]=\mathrm{NA}$;
$\operatorname{Project}$ passes that missing-value \mbox{indicator} and never indexes a
different topic family.
Available current topic keys select
\begin{equation}
\begin{aligned}
v_{\mathrm{fam}}(x)=\big(&p_{\mathrm{cb},T(x)}^{\mathrm{mer}}
[k_{\mathrm{mer},T(x)}(x)],\\[-2pt]
&p_{\mathrm{cb},T(x)}^{\mathrm{mcc}}[k_{\mathrm{mcc},T(x)}(x)],\\[-2pt]
&p_{\mathrm{cb},T(x)}^{\mathrm{loc}}[k_{\mathrm{loc},T(x)}(x)]\big).
\end{aligned}
\label{eq:card-behavior-projection}
\end{equation}
At scoring time, the stored summary remains fixed.

\paragraph{Illustrative activity shares.}
These values illustrate Eq.~\ref{eq:card-behavior-projection}, not observed
results. Suppose the card's shares for the resolved Merchant, MCC, and Location
topics are $v_{\mathrm{fam}}(x)=(0.05,0.08,0.30)$. If only the Merchant-topic
key is unresolved, the first entry is missing, while the MCC share of 0.08 still
fills the second entry.

\section{A Worked Persona-Protocol Example}
\label{sec:protocol-example}

This example uses the selected topic scheme to illustrate topic content and
hierarchy, and then shows the kind of revision that protocol search supports.
The sanitized travel-topic narrative in Section~\ref{sec:design} reports transaction
size, channel mix, cross-location activity, and approval behavior relative to
their respective overall baselines. These comparisons describe the topic-level
record~rather than specify membership. Separately, under one parent MCC topic,
the topic scheme contains Merchant topics whose labels distinguish predominantly
online domestic activity from
predominantly online cross-location activity within the same qualitative
transaction-size band. These labels describe behavioral clusters within the
parent MCC topic,
not a hand-written assignment rule.

\paragraph{An illustrative revision.}
Consider a candidate narrative schema $N$ that permits an interpretation such
as \emph{issuer caution is visible in lower approval}. A lower approval rate is
an observed comparison; it does not establish issuer intent. The Protocol Critic
could diagnose this explanation as unsupported.
The Narrative Schema role could then revise $N$ to require an observed
field-and-baseline comparison without attributing intent. The resulting
description would state \emph{approval is below its overall baseline}.
This would change the rule governing subsequent narratives, not just one
sentence; it would not by itself change topic membership.
A better-grounded description is not sufficient for acceptance. A candidate replaces
the current best protocol only if development BM1 improves and the prespecified
checks pass; a
source-consistent candidate without improvement is rejected, as is an
improving candidate that fails a required check. The critic's diagnosis guides
the next proposal but cannot override that rule.

\begin{table}[t]
\caption{Components of the persona protocol, with one instance of each: behavior
signals $F$, narrative schema $N$, topic construction $Q$, retrieval rules $R$,
and release checks $G$.}
\label{tab:protocol-components}
\centering
\footnotesize
\setlength{\tabcolsep}{3pt}
\renewcommand{\arraystretch}{1.0}
\begin{tabular}{@{}c>{\raggedright\arraybackslash}p{0.50\columnwidth}>{\raggedright\arraybackslash}p{0.35\columnwidth}@{}}
\toprule
Part & What it fixes & Example \\
\midrule
$F$ & Behavior signals, history windows, baselines, and support rules
    & cross-location share against a peer baseline \\
$N$ & Numerical-evidence coverage, narrative fields, and source-consistency checks
    & 4--6 selected behavior signals per narrative \\
$Q$ & Semantic topic definitions, Merchant subdivision, and topic identities
    & behavior-feature clustering within an MCC topic \\
$R$ & Exact keys, minimum support, fallback order, and missing values
    & low-support pair falls back to a predefined marginal, here MCC-only \\
$G$ & Schema, cutoff-validity, freshness, drift, and compatibility checks
    & a stale build leaves the last validated package active \\
\bottomrule
\end{tabular}
  \vspace{0.25pt}
\end{table}

\paragraph{After the protocol is frozen.}
Newly eligible history supplies updated numerical evidence, and the build
applies the fixed topic definitions and topic-assignment rules to construct
records and
maps. For a previously unseen Merchant
without a supported \mbox{mapping,} the predefined parent MCC-topic fallback supplies
the context, with fallback level and support passed to the adapter.

\section{Protocol, Refresh, and Serving Algorithms}
\label{sec:algorithms}

The following algorithms make the three stages in Fig.~\ref{fig:framework}
explicit. Algorithm~\ref{alg:design} searches for and freezes the protocol.
Algorithm~\ref{alg:refresh} applies that protocol on a schedule and activates a
compatible release package only after validation. Algorithm~\ref{alg:serve}
retrieves records from the active package and computes one deterministic score.
We denote detector~$m$'s adapter by $a_m$, its parameters by $\theta_m$, and the
metadata input of Section~\ref{sec:inference} by $z_{\mathrm{meta}}(x)$. During
design, an LLM operating under a shared configuration is prompted to perform the
predefined proposal and critic roles, and an approved open-weight generator
writes the grounded narratives under schema $N$.
\mbox{Algorithms}~\ref{alg:refresh} and~\ref{alg:serve} have no LLM calling.

\paragraph{Fixed search inputs.} The evaluation procedure $H$, selection rule
$g_{\mathrm{dev}}$, and search budget are fixed inputs, not outputs of an LLM
role. Each candidate's $G$ specifies its release checks; revisions to $G$ are
themselves evaluated under the same $H$ and $g_{\mathrm{dev}}$.
The execution configuration $E^\star$ records software and execution settings
separately from the selected protocol. It pins the generator and embedding
configuration, which are reproducible internally but \mbox{unnamed} under our
disclosure policy.

\begin{algorithm}[htbp]
\caption{Bounded persona-protocol self-evolution}
\label{alg:design}
\footnotesize
\begin{algorithmic}[1]
\Require Initial protocol $\Pi_0$, prespecified execution configuration $E$,
build data
$D_{\mathrm{build}}$, development set $D_{\mathrm{dev}}$
\Require Frozen detector $f_4$, search budget $J$, prespecified evaluation procedure
$H$, deterministic selection rule $g_{\mathrm{dev}}$, prespecified stopping rule
\State $\Pi_{\mathrm{best}}\gets\Pi_0$; $A\gets\emptyset$; $d\gets\emptyset$
\State $h_{\mathrm{best}}\gets H(\Pi_{\mathrm{best}},E,f_4;
D_{\mathrm{build}},D_{\mathrm{dev}},\mathrm{BM1})$
\For{$j=1,\ldots,J$}
  \State $\Pi_j\gets\mathrm{AssembleProtocolRevision}(\Pi_{\mathrm{best}},A,d)$
  \Comment{roles: $F,N,Q$; $d$ guides $R,G$}
  \State $h_j\gets H(\Pi_j,E,f_4;D_{\mathrm{build}},D_{\mathrm{dev}},\mathrm{BM1})$
  \Comment{rebuild memory, fit temporary adapter}
  \State $d\gets\mathrm{ProtocolCritic}(\Pi_j,h_j)$
  \State $\mathrm{accept}\gets g_{\mathrm{dev}}(h_j\mid h_{\mathrm{best}})$
  \State $\delta_\Pi\gets\mathrm{ProtocolDiff}(\Pi_j,\Pi_{\mathrm{best}})$
  \State append $(\delta_\Pi,h_j,d,\mathrm{accept})$ to $A$
  \If{$\mathrm{accept}=1$}
    \State $\Pi_{\mathrm{best}}\gets\Pi_j$; $h_{\mathrm{best}}\gets h_j$
  \EndIf
  \If{stopping rule holds} \State \textbf{break} \EndIf
\EndFor
\State $\Pi^\star\gets\Pi_{\mathrm{best}}$; $E^\star\gets E$
\State \Return frozen $(\Pi^\star,E^\star)$ and protocol-search log $A$
\end{algorithmic}
\end{algorithm}

For each candidate, $H$ rebuilds memory, fits a temporary adapter around frozen
Detector~4, and reports development BM1 and check results; the temporary
adapters are not released. The selection rule
decides acceptance; $\mathrm{ProtocolDiff}$ records the changed components.
The log retains proposals, diagnoses, and decisions.

\begin{algorithm}[htbp]
\caption{Scheduled offline memory refresh and release-package activation}
\label{alg:refresh}
\footnotesize
\begin{algorithmic}[1]
\Require Frozen $(\Pi^\star,E^\star)$, weekly cutoff $T$, last active package
$\mathcal P^{-}$, compatible adapters $\Theta$
\State load the selected topic definitions and identities from $\Pi^\star$
\State $D_T\gets\mathrm{EligiblePriorEvidence}(T)$
\State $\mathcal V_T\gets
\mathrm{BuildAllViews}(D_T,\mathcal L;F,Q,R,E^\star)$
\Comment{records, topic-assignment maps}
\State $K_T^{\mathrm{cand}}\gets
\mathrm{AssembleCompatibleSnapshot}(\mathcal V_T,T,\Pi^\star,E^\star)$
\State $M_T^{\mathrm{cand}}\gets
\mathrm{BuildReleaseMetadata}(K_T^{\mathrm{cand}},\Pi^\star,E^\star,\Theta)$
\State $\mathcal P_T^{\mathrm{cand}}\gets(K_T^{\mathrm{cand}},M_T^{\mathrm{cand}},\Theta)$
\State $\mathrm{status}\gets\mathrm{Validate}(\mathcal P_T^{\mathrm{cand}};G)$
\If{$\mathrm{status}=\mathrm{pass}$}
  \State $\mathcal P^{\mathrm{serve}}\gets\mathrm{AtomicActivate}(\mathcal P_T^{\mathrm{cand}})$
\Else
  \State $\mathcal P^{\mathrm{serve}}\gets\mathcal P^{-}$ \Comment{last known-good package}
\EndIf
\State \Return $\mathcal P^{\mathrm{serve}}$
\end{algorithmic}
\end{algorithm}

Each weekly run of Algorithm~\ref{alg:refresh} updates all five numerical views
under the selected topic
definitions; it does not repeat protocol search. The complete package is
activated in one switch, and a failed validation leaves the prior package active.

\begin{algorithm}[htbp]
\caption{Online resolved-key retrieval and deterministic scoring}
\label{alg:serve}
\footnotesize
\begin{algorithmic}[1]
\Require Transaction $x$ with event time $t(x)$, detector index $m$, frozen
detector $f_m$, registry $\mathcal A$ mapping event times to active compatible packages
\State $s_{\mathrm{base}}\gets f_m(x)$
\State $\mathcal P_x\gets\mathrm{PinActivePackage}(\mathcal A,t(x),m)$
\If{$\mathcal P_x$ unavailable or its health check fails}
  \State \Return $s_{\mathrm{base}}$ \Comment{safe exit to the base score}
\EndIf
\State extract $K_{T(x)}$, $\beta_{T(x)}$, $\theta_m$, and the topic scheme from $\mathcal P_x$
\State $\kappa_{T(x)}(x)\gets\mathrm{ResolveKeys}(K_{T(x)},x)$
\If{key-resolution service or schema validation fails}
  \State \Return $s_{\mathrm{base}}$ \Comment{service failure; a sparse or unseen key is a normal fallback}
\EndIf
\ForAll{$\ell\in\{\mathrm{cb,mer,mcc,loc,int}\}$}
  \State retrieve $r_\ell(x)$ under Eq.~\ref{eq:lookup}
  \If{infrastructure or schema/interface failure} \State \Return $s_{\mathrm{base}}$ \EndIf
\EndFor
\State derive $\phi(x)$ under Eq.~\ref{eq:retrieve}
\If{feature-vector construction fails} \State \Return $s_{\mathrm{base}}$ \EndIf
\State $(z_{\mathrm{meta}}(x),\mathrm{status})\gets
\mathrm{DeriveMetadata}(x,K_{T(x)},\kappa_{T(x)}(x))$
\If{$\mathrm{status}=\mathrm{fail}$} \State \Return $s_{\mathrm{base}}$ \EndIf
\State $(s_{\mathrm{final}},\mathrm{status})\gets
\mathrm{ExecuteAdapter}(a_m,s_{\mathrm{base}},
\phi(x),z_{\mathrm{meta}}(x))$
\If{$\mathrm{status}=\mathrm{fail}$} \State \Return $s_{\mathrm{base}}$ \EndIf
\State \Return $s_{\mathrm{final}}$
\end{algorithmic}
\end{algorithm}

The serving procedure (Algorithm~\ref{alg:serve}) uses explicit safe exits so
scoring can continue when PersonaAgent's serving path fails. If package
pinning, key resolution, record retrieval, feature-vector construction,
metadata derivation, or adapter execution fails, the transaction retains the
frozen detector's original score.

\section{Supplementary Diagnostics}

\paragraph{Confidence intervals and refits.}
\label{sec:uncertainty}
The intervals in Table~\ref{tab:business-results} come from a paired bootstrap
over transactions. Within each period, each of 1{,}000 draws resamples
transactions with replacement, keeping each transaction's label and its base and
final scores together, and recomputes every relative improvement; the 2.5th and 97.5th percentiles
of the draws give the interval. The same draws serve all four detectors and all three
metrics within a period. To measure fitting variability, we refit the fusion
network with five seeds while keeping persona records, view encoders, base
scores, and data fixed. All 24 comparisons show improvements in every refit, and the
standard deviation of the relative improvement across seeds, averaged over the eight detector--period
cells, is 1.21 (BM1), 0.15 (PR-AUC), and 0.013 (AUC) percentage points.

\paragraph{Cross-period performance.}
\label{sec:period-analysis}
Both periods evaluate the same adapters, frozen before Period~1. From Period~1
to Period~2, the relative improvement in BM1 decreases by approximately two-fifths for Detector~1
and nearly three-tenths for Detector~2, while it more than doubles for Detector~3
(Table~\ref{tab:business-results}). The gains vary across detectors and
metrics but remain positive in Period~2.

\paragraph{Record age.}
\label{sec:record-age}
To test whether fresher records matter, we rescore Period~2 after replacing
only the frozen package's MCC-topic, Location-topic, and interaction records
with those of the preceding weekly build. The protocol, encoders, and adapters
remain fixed within each comparison. The package's own, fresher records give
a higher relative improvement in BM1 for every detector, by 0.86, 0.78, 2.52, and 0.82 percentage points averaged over seeds for Detectors~1--4, and the difference is positive in every seed. The relative improvement in PR-AUC is higher by 0.16, 0.25, and 0.28 percentage points for Detectors~1--3
and unchanged for Detector~4; the relative improvement in AUC changes by at most 0.02 percentage points.

\balance
\paragraph{Matched ablations.}
\label{sec:matched-controls}
The matched ablations in Section~\ref{sec:rq2} refit the fusion network on the
same training rows with the same architecture and fitting procedure, changing
only its inputs: the base score alone; the base score and the metadata-adjusted
base score;~or these two scores with the five view predictions, as in
Eq.~\ref{eq:fusion-input}.
Variants with views reuse the same out-of-fold encoder predictions. Each variant
scores the transactions of both periods offline, and reported values average
three random seeds; differences between variants are in percentage points of
relative improvement.

\paragraph{Correction-fingerprint diagnostic.}
\label{sec:correction-fingerprint}

We define a \emph{correction fingerprint} for each detector and period
as a seven-dimensional effect-size vector contrasting recovered anomalies with
lost detections. Recovered anomalies are confirmed anomalies flagged
only by PersonaAgent; lost detections are confirmed anomalies flagged only by
the frozen detector. Both cohorts use the retrospective analysis thresholds of
Section~\ref{sec:deployment}. The seven features are the five persona-view scores,
transaction amount, and Merchant-topic-relative amount. For each feature,
Cliff's delta is the fraction of recovered/lost pairs where the recovered
transaction's value is higher minus the fraction where it is lower. Missing
values are omitted separately by feature. These seven signed effects form the
correction fingerprint, and its cosine similarity across periods is near 1 when
they keep the same signs and relative sizes in both periods.

\end{document}